\documentclass[acmsmall,screen]{acmart}
\usepackage{xcolor}
\usepackage{graphicx}
\usepackage{multirow}
\usepackage{fontawesome5} 
\usepackage[many]{tcolorbox} 
\usepackage{subcaption} 
\usepackage{pifont}
\usepackage{booktabs}
\usepackage{makecell}
\usepackage{arydshln}
\usepackage{enumitem}
\definecolor{ForestGreen}{RGB}{64, 136, 39}
\definecolor{redish}{RGB}{212, 57, 57}
\newtcolorbox{boxK}{
    top=2pt,
    bottom=2pt,
    left=2pt,
    right=2pt,
    boxrule = 0pt,
    toprule = 0pt, 
}
\makeatletter
\DeclareRobustCommand\onedot{\futurelet\@let@token\@onedot}
\def\@onedot{\ifx\@let@token.\else.\null\fi\xspace}

\makeatother
\AtBeginDocument{
  }
\begin{document}
\title{Evaluating the Formal Reasoning Capabilities of Large Language Models through Chomsky Hierarchy}
\author{Yihong Dong}
\email{dongyh@stu.pku.edu.cn}
\affiliation{
  \institution{Peking University}
  \city{Beijing}
  \country{China}
}
\author{Jianha Xiao}
\email{hamiltonxiao@stu.pku.edu.cn}
\affiliation{
  \institution{Peking University}
  \city{Beijing}
  \country{China}
}
\author{Xue Jiang}
\email{jiangxue@stu.pku.edu.cn}
\affiliation{
  \institution{Peking University}
  \city{Beijing}
  \country{China}
}
\author{Xuyuan Guo}
\email{xyguo25@stu.pku.edu.cn}
\affiliation{
  \institution{Peking University}
  \city{Beijing}
  \country{China}
}
\author{Zhiyuan Fan}
\email{zyfan043@gmail.com}
\affiliation{
  \institution{Peking University}
  \city{Beijing}
  \country{China}
}
\author{Jiaru Qian}
\email{qianjiaru77@gmail.com}
\affiliation{
  \institution{Peking University}
  \city{Beijing}
  \country{China}
}
\author{Kechi Zhang}
\email{zhangkechi@pku.edu.cn}
\affiliation{
  \institution{Peking University}
  \city{Beijing}
  \country{China}
}
\author{Jia Li}
\email{jia.li@whu.edu.cn}
\affiliation{
  \institution{Wuhan University}
  \city{Wuhan}
  \country{China}
}
\author{Zhi Jin}
\email{zhijin@pku.edu.cn}
\affiliation{
  \institution{Peking University}
  \city{Beijing}
  \country{China}
}
\author{Ge Li}
\email{lige@pku.edu.cn}
\affiliation{
  \institution{Peking University}
  \city{Beijing}
  \country{China}
}

\renewcommand{\shortauthors}{Dong et al.}

\begin{abstract}
The formal reasoning capabilities of Large Language Models (LLMs) are crucial for advancing automated software engineering. However, existing benchmarks for LLMs lack systematic evaluation based on computation and complexity, leaving a critical gap in understanding their formal reasoning capabilities. Therefore, it is still unknown whether state-of-the-art (SOTA) LLMs can grasp the structured, hierarchical complexity of formal languages as defined by Theory of Computation. To address this, we introduce ChomskyBench, a benchmark for systematically evaluating LLMs through the lens of the Chomsky Hierarchy. Unlike prior work that uses vectorized classification for neural networks, ChomskyBench is the first to combine \textbf{full Chomsky Hierarchy coverage}, \textbf{process-trace evaluation via natural language}, and \textbf{deterministic symbolic verifiability}. ChomskyBench is composed of a comprehensive suite of language recognition and generation tasks designed to test capabilities at each level: regular (Type-3), deterministic/non-deterministic context-free (Type-2), context-sensitive (Type-1), and recursively enumerable (Type-0). 
Extensive experiments on SOTA LLMs using ChomskyBench demonstrate a clear performance stratification that correlates with the hierarchy's levels of complexity. Our analysis reveals a direct relationship where increasing task difficulty substantially impacts both inference length and performance. Furthermore, we find that while larger models and advanced inference methods offer notable relative gains, they face severe efficiency barriers: achieving practical reliability would require prohibitive computational costs, revealing that current limitations stem from inefficiency rather than absolute capability bounds. A time complexity analysis further indicates that LLMs are significantly less efficient than traditional algorithmic programs for these formal tasks. These results delineate the practical limits of current LLMs, highlight the indispensability of traditional software tools, and provide insights to guide the development of future LLMs with more powerful formal reasoning capabilities.\footnote{The dataset and code are available at \url{https://github.com/stackupdown/ChomskyBench}.}
\end{abstract}
\maketitle
\section{Introduction}
Large Language Models (LLMs) have emerged as a transformative force in software engineering, demonstrating remarkable capabilities in tasks ranging from code generation and completion to bug fixing and documentation \cite{hou2024large, jiang2024survey, mu2024clarifygpt, zhou2024out, dvivedi2024comparative,Self-Collaboration,Self-Planning,Agent4code,ROCODE}\nocite{SEED, Saber}. Their proficiency in processing and generating code has catalyzed a new wave of research in Automated Software Engineering, promising to enhance developer productivity and streamline complex development workflows. However, the impressive performance of these models often masks a critical and unanswered question: Do LLMs possess genuine formal reasoning capabilities, or are they merely sophisticated pattern matchers? 
Resolving this question is essential for ensuring LLMs' reliability in software engineering domains that necessitate strict adherence to formal rules, such as compiler design \cite{cummins2025llm}, protocol verification \cite{mao2025llm}, static analysis \cite{li2024enhancing}, and so on.
\begin{figure}[t]
    \centering
    \includegraphics[width=0.98\linewidth]{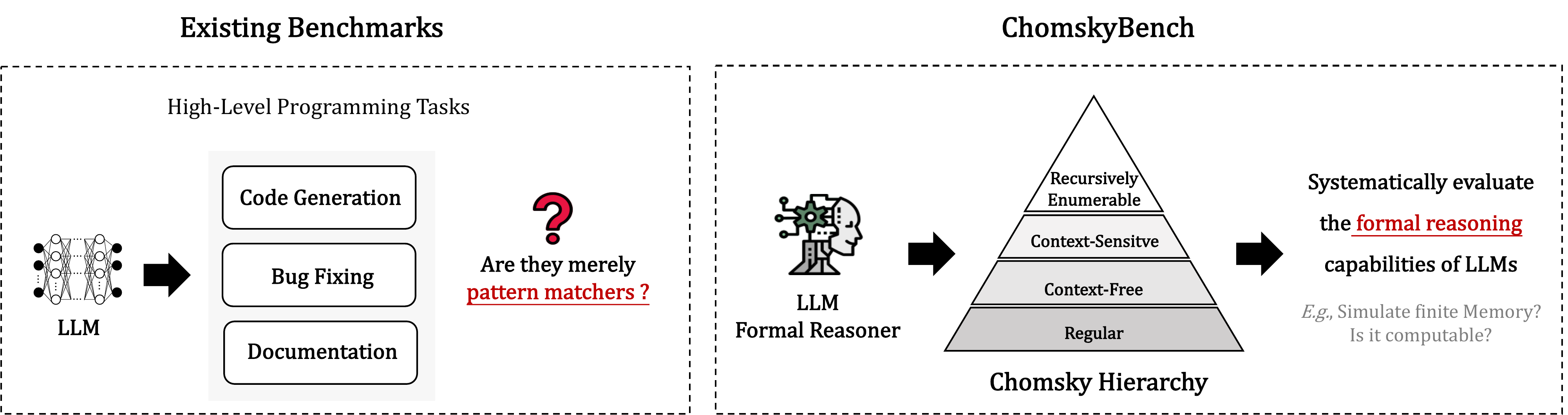}
    \caption{A conceptual comparison between existing benchmarks and our proposed benchmark.}
    \label{fig:motivation}
\end{figure}
Despite a plethora of benchmarks designed to evaluate LLMs on practical programming tasks in software engineering \cite{humaneval, livecodebench, li2024devevalmanuallyannotatedcodegeneration, li2024evocodebenchevolvingcodegeneration}, a fundamental gap persists in the understanding of their core computational abilities. Existing evaluations, which focus on the functional correctness of code in various programming languages (e.g., Python, Java, C++), are useful but conflate syntactic and semantic understanding with the underlying logic of formal systems. They lack a systematic, principled framework grounded in the Theory of Computation to probe the limits of LLM reasoning, as illustrated in Figure \ref{fig:motivation}. Consequently, it remains unclear whether LLMs can fundamentally grasp the hierarchical complexity inherent in formal languages, a cornerstone of theoretical computer science established by Noam Chomsky \cite{chomsky1956three}. Without such a foundational understanding, we risk building the future of automated software development on an unstable foundation, where the reasoning failures of LLMs are unpredictable and potentially catastrophic.
To bridge this gap, we introduce ChomskyBench, a novel benchmark designed to systematically evaluate LLMs through the lens of the Chomsky Hierarchy. This hierarchy provides a canonical, tiered framework for classifying formal languages based on their complexity, making it suitable for assessing the formal reasoning capabilities of LLMs. ChomskyBench comprises a comprehensive suite of language recognition and generation tasks designed to test formal reasoning capabilities at 4 levels: regular reasoning (Type-3), deterministic/non-deterministic context-free reasoning (Type-2), context-sensitive reasoning (Type-1), and recursively enumerable reasoning (Type-0). This hierarchical structure allows us to not only evaluate the final correctness of a model's output but also to assess its fundamental ability to handle concepts like recursion, nesting, and long-range dependencies, pinpointing the specific breaking points in its reasoning abilities. 
Leveraging ChomskyBench, we conduct an extensive empirical study on a range of SOTA LLMs, including both closed-source and open-source models. Our findings are striking: LLM performance directly correlates with the Chomsky Hierarchy, exhibiting a significant and consistent degradation as the complexity of the formal language increases. While larger LLMs and advanced prompting techniques, \emph{e.g.}, Chain-of-Thought (CoT), provide notable relative improvements, achieving practical reliability would require prohibitive computational costs \cite{COT,CodeRL+,RL-PLUS,Think-Anywhere}. Furthermore, a time complexity analysis reveals that even on tasks where they succeed, LLMs are orders of magnitude less efficient than simple, traditional algorithms. Analysis of failure cases reveals a core inability to manage deep recursion and maintain state over long dependencies, which becomes more prominent at higher levels of the hierarchy.
The main contributions of this work are highlighted as follows:
\begin{itemize}
    \item We propose the Chomsky Hierarchy as a systematic framework for evaluating the formal reasoning capabilities of LLMs, providing a principled theoretical foundation for diagnosing their computational limits.
    \item We introduce ChomskyBench, the first benchmark offering full Chomsky Hierarchy coverage (including Type-0), process-trace evaluation via natural language, and deterministic symbolic verifiability.
    \item We provide extensive experiments showing that LLMs face severe efficiency barriers across Chomsky Hierarchy. While test-time scaling shows consistent improvements with no hard ceiling, achieving practical reliability requires prohibitive computational costs.
    \item Our findings delineate the theoretical limitations of SOTA LLMs, underscore the indispensability of traditional software tools, and offer insights to guide the development of LLMs with more powerful formal reasoning capabilities.
\end{itemize}

\section{Background and Related Work}
\noindent\paragraph{Scope of Formal Reasoning in This Work}
We focus on \textbf{computational/procedural formal reasoning}, i.e., the ability to simulate algorithmic processes (state transitions, stack manipulations, tape operations) defined by abstract automata. This is distinct from semantic, causal, or theorem-proving reasoning. By isolating foundational computational primitives, we enable precise diagnosis of where LLMs' mechanisms break down.
Our research is situated at the intersection of three key domains: the evaluation of LLMs on software engineering tasks, the investigation of their fundamental reasoning abilities and theoretical limits, and the landscape of benchmarks for formal reasoning. This section reviews prior work in these areas to contextualize the unique contribution of ChomskyBench.
\begin{figure}[t]
    \centering
    \includegraphics[width=0.98\linewidth]{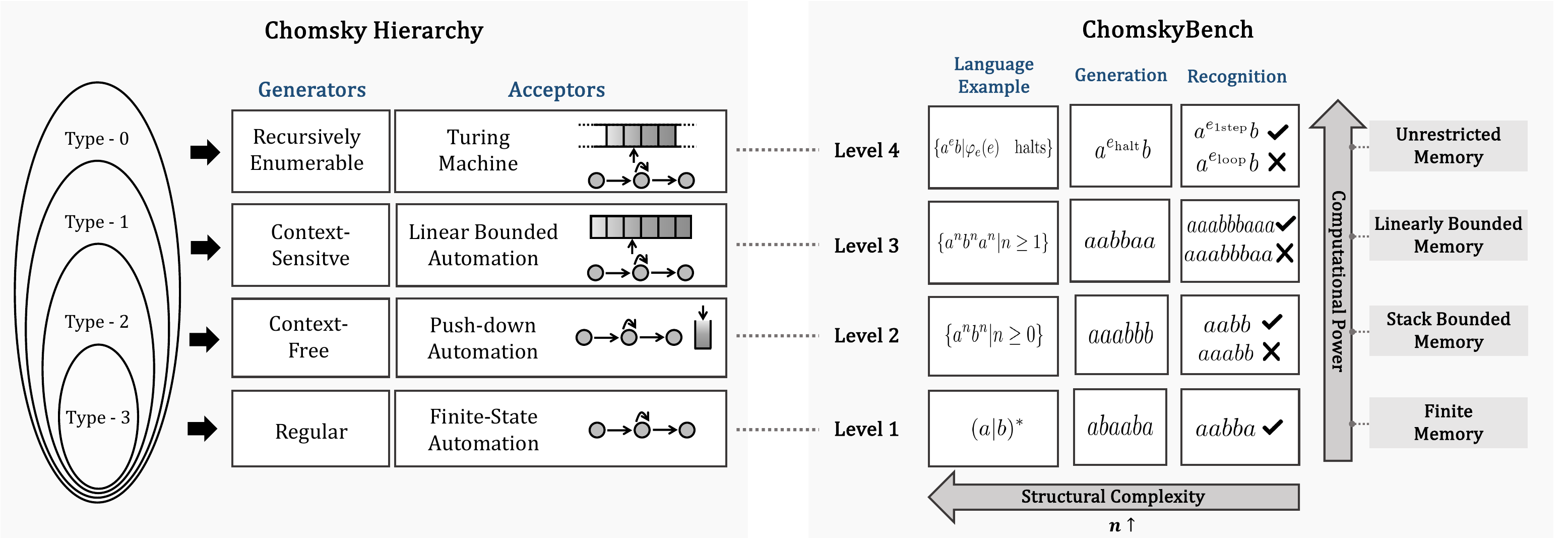}
    \caption{Chomsky Hierarchy and ChomskyBench: Four computational complexity levels (Type-3 to Type-0) with corresponding automata models and representative formal languages used for systematic evaluation of LLM formal reasoning capabilities.}
    \label{fig:chomsky_hierarchy}
\end{figure}
\subsection{LLM Evaluation in Software Engineering}
A vast body of literature has focused on evaluating the proficiency of LLMs in software engineering contexts. Some famous benchmarks, such as HumanEval \cite{humaneval}, MBPP \cite{MBPP}, and APPs \cite{APPS}, mainly focus on measuring the functional correctness of model-generated code, as well as their extended test cases version, i.e., HumanEval-ET, MBPP-ET, and APPs-ET \cite{Codescore}. More recent benchmarks extend this paradigm in several directions. For instance, LiveCodeBench \cite{livecodebench} emphasizes contamination-free evaluation by continuously collecting new problems from competitive programming platforms and expanding the scope of assessment to capabilities such as self-repair, code execution, and test output prediction. SWE-bench \cite{jimenez2024swebenchlanguagemodelsresolve} shifts the focus to real-world maintenance tasks, requiring models to resolve GitHub issues by editing large, practical codebases. Similarly, DevEval \cite{li2024devevalmanuallyannotatedcodegeneration} and EvoCodeBench \cite{li2024evocodebenchevolvingcodegeneration} build on realistic repository-level distributions, offering fine-grained annotations, dependency-aware settings, and evolving datasets to mitigate data leakage, thereby capturing a broader spectrum of developer-oriented tasks. To address the limitations of traditional benchmarks that only evaluate pre-existing knowledge, KOCO-Bench \cite{KOCO-BENCH} provides a companion knowledge base alongside its test set. This enables the evaluation of a model's ability to learn and apply new domain knowledge (e.g., APIs, rules, and constraints).
While invaluable for assessing practical coding aptitude, this dominant evaluation paradigm has a fundamental limitation: its primary focus is on the outcome of coding tasks (i.e., does the code work?) rather than the underlying computational reasoning. This approach inherently conflates knowledge of language-specific APIs, algorithmic recall, and genuine formal reasoning, making it difficult to isolate and analyze a model's core computational capabilities. Our work diverges from this by abstracting away from specific programming languages to test the foundational principles of formal language recognition and generation, which are foundational to all programming.
\subsection{Probing the Reasoning Limits and Theoretical Power of LLMs}
Concurrently with empirical evaluations, a significant thread of research has investigated the theoretical computational power of the base architecture of LLMs, i.e., Transformer \cite{vaswani2023attentionneed}. Several studies \cite{bhattamishra2020computationalpowertransformersimplications,attentionisturing,li2025constantbitsizetransformersturing} have argued that, with sufficient scale and modifications like recurrence or memory, Transformers are Turing-complete in principle. 
However, other theoretical work \cite{rizvi2024simulatingweightedautomatasequences, zhang2025finitestateautomatainside} highlights that standard, finite-precision Transformers are formally equivalent to finite-state automata. This implies they cannot, in theory, recognize languages requiring unbounded memory or recursion, such as many context-free or context-sensitive languages.
This creates a critical tension between the theoretical potential of LLMs and their practical, architectural limitations. Our work provides an empirical bridge for this theoretical debate. By systematically testing where current LLMs actually fail on a graded spectrum of computational complexity, ChomskyBench is designed to offer concrete evidence of these practical limitations, mapping the effective reasoning power of LLMs against the canonical yardstick of the Chomsky Hierarchy.
\begin{table}[h]
  \caption{Difference of ChomskyBench and previous formal language benchmark, where each task of ChomskyBench can change the structural complexity by modifying the input length of strings.}
  \resizebox{\textwidth}{!}{
  \begin{tabular}{lcccccccc}
    \toprule
    \multirow{2}{*}{\textbf{Dataset}} & \multicolumn{5}{c}{\textbf{Chomsky Level (Task Num)}} & \multirow{2}{*}{\textbf{Total}} & \multirow{2}{*}{\textbf{Question Focus}} & \multirow{2}{*}{\textbf{I/O Type}} \\
    \cmidrule(lr){2-6}
    & RE & CS & NCF & DCF & Regular & & & \\
    \midrule
    \multirow{2}{*}{NNCH \cite{deletang2022neural}} & \textcolor{redish}{\ding{56}} & \textcolor{teal}{\ding{52}} & \textcolor{redish}{\ding{56}} & \textcolor{teal}{\ding{52}} & \textcolor{teal}{\ding{52}} & \multirow{2}{*}{15} & \multirow{2}{*}{Final state} & \multirow{2}{*}{One-hot Vectors} \\
    & \textcolor{redish}{0} & 7 & \textcolor{redish}{0} & 4  & 4  & & & \\
    \hdashline
    \multirow{2}{*}{FLaRe \cite{butoi2024training}} & \textcolor{redish}{\ding{56}} & \textcolor{teal}{\ding{52}} & \textcolor{teal}{\ding{52}} & \textcolor{teal}{\ding{52}} & \textcolor{teal}{\ding{52}} & \multirow{2}{*}{18} & \multirow{2}{*}{Final state} & \multirow{2}{*}{One-hot Vectors} \\
    & \textcolor{redish}{0} & 7 & 3 & 1 & 7 & & & \\
    \hdashline
    \multirow{2}{*}{\textbf{ChomskyBench (Ours)}} & \textcolor{teal}{\textbf{\ding{52}}} & \textcolor{teal}{\textbf{\ding{52}}} & \textcolor{teal}{\textbf{\ding{52}}} & \textcolor{teal}{\textbf{\ding{52}}} & \textcolor{teal}{\textbf{\ding{52}}} & \multirow{2}{*}{\textbf{115}} & \multirow{2}{*}{\textbf{\makecell{Process Trace \\ + Final state}}} & \multirow{2}{*}{\textbf{\makecell{Natural Language}}} \\
    & \textbf{20} & \textbf{27} & \textbf{15} & \textbf{29} & \textbf{24} & & & \\
    \bottomrule
  \end{tabular}}
  \label{tab:datasets}
\end{table}
\subsection{Benchmarks for Formal Language and Reasoning}
The research closely related to our work falls into two categories: studies on specific formal language classes and benchmarks for formalized reasoning.
A significant line of prior investigation has explored the ability of neural architectures to learn formal languages \cite{wu2022autoformalizationlargelanguagemodels, xia2025largelanguagemodelslearn, wang2025kiminaproverpreviewlargeformal, ganguly2024proofthoughtneurosymbolic, shao2024deepseekmathpushinglimitsmathematical,soroco2025pdecontrollerllmsautoformalizationreasoning}. While foundational, these studies are often fragmented, focusing on a narrow slice of the Chomsky Hierarchy, such as the boundary between regular and context-free languages. More benchmarks, such as NNCH \cite{deletang2022neural} and FLaRe \cite{butoi2024training}, have emerged, but they suffer from two limitations that motivate our work: 1) Their hierarchical coverage is incomplete. NNCH lacks tasks for NCF tasks, which are essential for testing a model's ability to handle ambiguity. Both NNCH and FLaRe completely exclude recursively enumerable tasks, which are necessary to probe the theoretical limits of computation corresponding to Turing machines. 2) Their evaluation method is misaligned with the reasoning capabilities of LLMs. These benchmarks are designed to test traditional neural networks, framing language recognition as a direct string classification problem. Consequently, their tasks rely on vectorized inputs and outputs, testing a network's ability to learn a mapping from an input sequence to a final state vector. In contrast, ChomskyBench leverages the natural language interface of LLMs to demand explicit computational reasoning, requiring not just a final state but a verifiable trace of the abstract machine's execution (e.g., stack manipulations or tape movements). This shift from outcome to process enables a more rigorous assessment of genuine formal reasoning.
A second, recent body of work focuses on complex reasoning within formalized systems. One major track targets automated theorem proving, with benchmarks like MiniF2F (Olympiad-level math problems) \cite{zheng2022minif2fcrosssystembenchmarkformal}, ProofNet (undergraduate-level theorems) \cite{azerbayev2023proofnetautoformalizingformallyproving}, LeanEuclid (classical geometry proofs) \cite{murphy2024autoformalizingeuclideangeometry}, and large-scale datasets such as LeanDojo \cite{yang2023leandojotheoremprovingretrievalaugmented} and Lean Workbook \cite{ying2025leanworkbooklargescalelean}, which extend Lean-based evaluation by providing tens of thousands of formal-informal problem pairs and proof libraries. Other resources, such as PDA (Form4) \cite{pda} and MMA \cite{mma}, further broaden this landscape, emphasizing autoformalization of natural language statements into Lean 4, or bidirectional translation between formal and informal mathematics across multiple languages and domains. Complementary efforts also emphasize formal specification and verification from natural language, such as nl2spec \cite{cosler2023nl2specinteractivelytranslatingunstructured}, which translates requirements into temporal logic. These studies highlight the increasing sophistication of benchmarks that test the ability of LLMs to handle structured mathematical reasoning, formal verification, and proof generation.
However, a critical gap remains. To our knowledge, no prior work has leveraged the full Chomsky Hierarchy as a systematic and comprehensive framework for evaluating the foundational computational power of LLMs. The first line of work is too fragmented or methodologically misaligned, while the second line, focused on advanced theorem proving, presupposes the computational competencies that have yet to be systematically measured. ChomskyBench addresses this lacuna. It is the first benchmark to provide a unified suite of tasks spanning all four levels of the hierarchy, from Type-3 to Type-0. 
This holistic approach allows us to map LLM formal reasoning capabilities and delineate the theoretical boundaries of their architectural limitations.

\section{ChomskyBench}
To conduct a systematic and diagnostic evaluation of the formal reasoning capabilities of LLMs, we design ChomskyBench, a benchmark framework deeply rooted in automata theory and formal languages. This section elaborates on ChomskyBench's foundational philosophy, design principles, detailed composition, and its meticulous construction and validation protocol, ultimately demonstrating its contribution and superiority.
\subsection{Motivation and Foundational Philosophy}
Current benchmarks for evaluating the coding abilities of LLMs, such as HumanEval \cite{humaneval}, LiveCodeBench \cite{livecodebench}, and SWE-bench \cite{jimenez2024swebenchlanguagemodelsresolve}, primarily focus on functional correctness. They employ unit tests to verify whether a model-generated code snippet produces the expected output for specific inputs. While valuable for measuring programming skills, it has two fundamental limitations:
\begin{itemize}
    \item \textbf{Conflation of Reasoning and Memorization:} Tasks in high-level programming languages inherently entangle underlying logical reasoning with the memorization of vast API libraries, specific syntactic sugar, and common programming paradigms. An LLM might "solve" a problem merely because it has encountered a similar solution in its pre-training corpus \cite{bubeck2023sparks}, which fails to prove its understanding of the underlying computational process.
    \item \textbf{Lack of Diagnostic Capability and "Black-Box" Evaluation:} When a model fails, a unit-test-based evaluation provides only a binary (pass/fail) outcome. It cannot reveal the root cause of the failure—was the model unable to handle deep recursion? Track long-range dependencies? Or manage the state of multiple variables simultaneously? This "know-what-but-not-why" evaluation model offers limited insight for guiding the architectural design and algorithmic improvement of next-generation models.
\end{itemize}
To overcome these limitations, we shift the evaluation from measuring application-level \textbf{functional correctness} to probing theoretical-level \textbf{formal computability}. We cease to ask, "Can the model write a sorting algorithm?" and instead ask, "Can the model simulate the most fundamental computational primitives required to implement a sorting algorithm?"
To this end, we introduce \textbf{ChomskyBench}. We select the cornerstone of theoretical computer science, the \textbf{Chomsky Hierarchy}, as the theoretical backbone of our evaluation framework. This hierarchy provides a universally accepted "measuring stick" for computational complexity, defining a stratified ladder of computational power (Finite Memory → Stack-based Memory → Linearly Bounded Memory → Unrestricted Memory). With ChomskyBench, we aim to measure the formal reasoning boundaries of current LLMs.
\begin{figure}[t]
    \centering
    \includegraphics[width=0.98\linewidth]{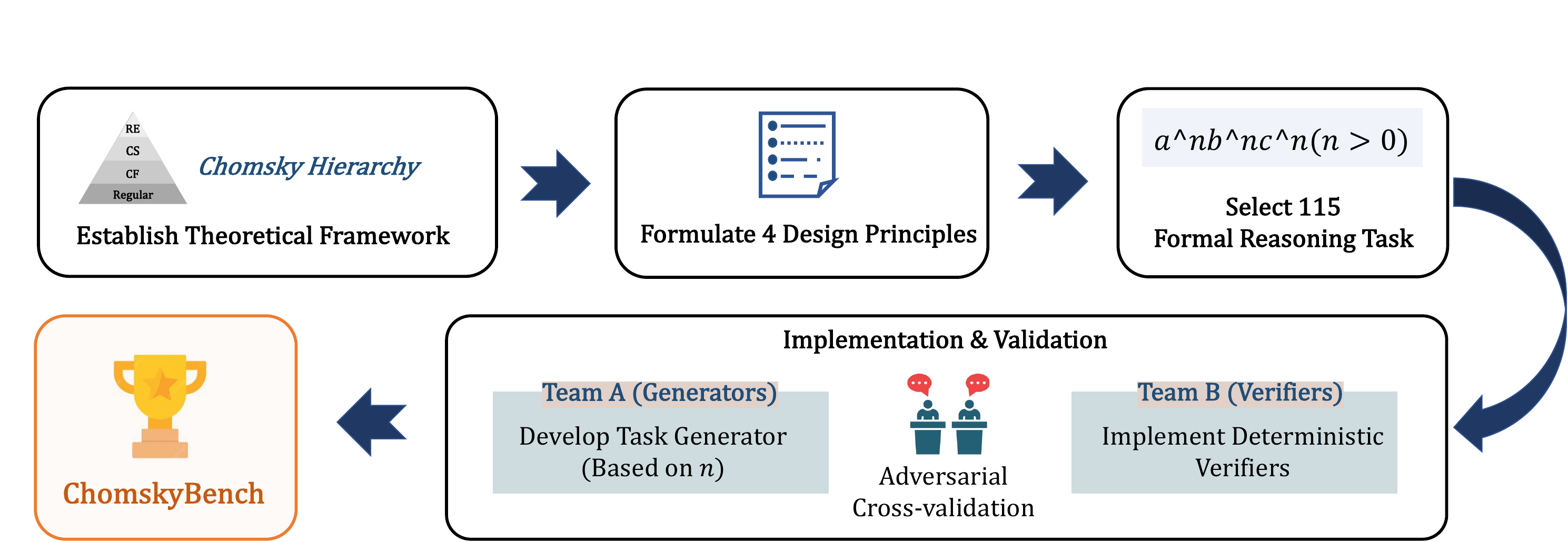}
    \caption{Construction Process of the ChomskyBench.}
    \label{fig:construction_pipline}
\end{figure}
\subsection{Design Principles}
The construction of ChomskyBench strictly adheres to four core principles.
\subsubsection{Principle 1: Hierarchical Abstraction for Diagnostic Analysis}
The core structure of ChomskyBench is aligned with all levels of the Chomsky Hierarchy (Type-3, Type-2, Type-1, and Type-0). Each level corresponds to a well-defined class of computational power and the theoretical automaton required to recognize it (Finite State Automata, Pushdown Automata, Linear-Bounded Automata, and Turing Machines, respectively). This hierarchical structure enables \textbf{differential diagnosis}: if LLMs excel at a Chomsky Hierarchy level tasks but their performance plummets on another Chomsky Hierarchy level tasks, we can draw a strong conclusion that the LLMs face a bottleneck in handling computations that require more than their Chomsky Hierarchy level. 
\subsubsection{Principle 2: Task Duality for a Holistic Assessment: Recognition and Generation}
To paint a comprehensive portrait of a model's reasoning profile, we design two complementary task paradigms for each formal language in the hierarchy:
\begin{itemize}
    \item \textbf{Recognition (Analytical Reasoning):} A decision problem where the model must determine if a given string belongs to a formal language (e.g., "Does the string `aaabbb' belong to the language $L = \{a^n b^n \mid n > 0\}$?"). This primarily tests the model's \textbf{analytical and deductive ability} to understand and apply a given set of rules to validate an input. 
    \item \textbf{Generation (Synthetic Reasoning):} A construction problem where the model must produce a novel, valid instance of a language according to its definition (e.g., "Generate a valid string from language $L = \{a^n b^n\}$ where $n=5$."). This mainly tests model's \textbf{synthetic and constructive ability} to internalize a set of rules and actively create a valid artifact.
\end{itemize}
Evaluating both capabilities provides a more complete picture of whether the model is merely passively matching rules or is capable of actively operationalizing them.
\subsubsection{Principle 3: Controlled Minimalism to Isolate Core Reasoning}
To rigorously evaluate LLMs' algorithmic reasoning, we commit to methodological reductionism \cite{nagel1979structure}. All tasks use minimalist alphabets (e.g., \{a,b,c\}, \{0,1\}) and unambiguous formal grammars, stripping away semantic content to compel direct engagement with structural and logical properties. Task difficulty is scaled along a parameterized complexity $n$ (e.g., string length, recursion depth), transforming evaluation into a controlled stress test that identifies the threshold where formal reasoning collapses.
\subsubsection{Principle 4: Absolute Algorithmic Verifiability}
Every task is grounded in a formal specification with deterministic, algorithmic verifiability. Model outputs are validated against deterministic verifiers (e.g., parsers, automata) that perfectly implement the formal language definition, guaranteeing mathematical certainty in correctness determination. This marks a departure from unit-test-based evaluation, ensuring fully automated, instantaneous, scalable, and reproducible assessment.
\subsection{Detailed Benchmark Architecture and Composition}
ChomskyBench is architected as a multi-level suite of tasks, with each level corresponding to a distinct tier of the Chomsky Hierarchy. This structure allows us to systematically probe a model's computational capabilities, from finite-state memory to universal algorithmic simulation. The languages chosen are canonical exemplars from theoretical computer science, selected for their unambiguous definitions and their direct correspondence to specific computational mechanisms.
\subsubsection{Systematic Data Generation}
To ensure a rigorous and reproducible evaluation, all data instances within ChomskyBench are generated through a programmatic, parameterized process.
Parameterized Complexity: Every language is defined with a complexity parameter $n$ (e.g., sequence length, counting depth). For our experiments, $n$ is varied across a range [2, $+\infty$) to create tasks of increasing difficulty, enabling us to identify each model's performance threshold.
\paragraph{Level 1: Type-3 (Regular Languages)}
Tests the ability to maintain finite state while processing sequences using Finite State Automata (FSA), probing basic pattern-matching capabilities.
\paragraph{Level 2: Type-2 (Context-Free Languages)}
Tests stack-based recursive processing using Pushdown Automata (PDA), stressing the capacity of memory and relating distant tokens.
\paragraph{Level 3: Type-1 (Context-Sensitive Languages)}
Tests multi-dependency tracking using Linear-Bounded Automata (LBA), challenging the model's ability to maintain and correlate multiple independent counts across long sequences.
\paragraph{Level 4: Type-0 (Recursively Enumerable Languages)}
Tests universal algorithmic simulation using Turing Machines, requiring the model to parse and trace step-by-step execution of computational processes. The ultimate test of whether an LLM can function as a general-purpose reasoning engine.
\subsection{Construction Protocol: Ensuring Scientific Rigor and Reproducibility}
\begin{figure}[t]
    \centering
    \includegraphics[width=0.8\linewidth]{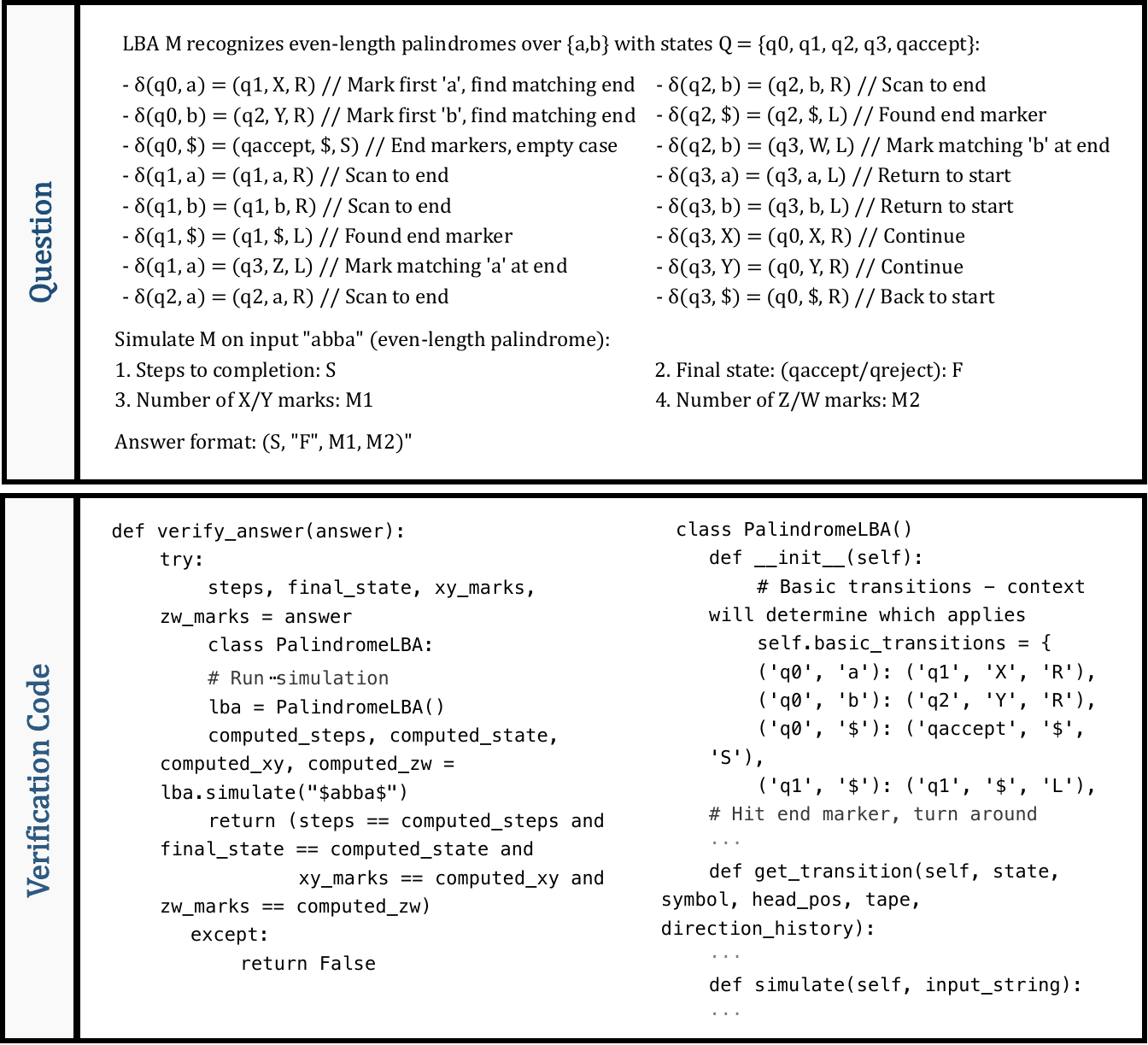}
    \caption{An Example in ChomskyBench.}
    \label{fig: example}
\end{figure}

\paragraph{Task Sources and Selection Criteria}
The 115 unique tasks in ChomskyBench are derived from two complementary sources: (1) \textbf{Classical problems from established textbooks} (approximately 15\%), drawn from canonical references such as ``Introduction to Automata Theory, Languages, and Computation'' \cite{hopcroft2001introduction} and ``Formal Languages and Their Relation to Automata' \cite{hopcroft1969formal}; and (2) \textbf{Novel problems designed by human experts} (approximately 85\%), formulated by our research team to ensure comprehensive coverage and resistance to data contamination. Tasks were selected to ensure balanced representation across all Chomsky Hierarchy levels.
\paragraph{Adversarial Cross-Validation Process}
The ChomskyBench benchmark is designed over approximately 860 person-hours by a team of five researchers with expertise spanning formal languages, computational theory, and compiler technology. To guarantee its absolute correctness, we employ an \textbf{adversarial cross-validation process}: the team is divided into two groups. One group (Team A) is responsible for designing the formal language specifications and procedural test-case generation scripts. The other group (Team B) independently implements traditional, algorithmically deterministic verifiers for each task (e.g., FSA simulators, PDA parsers). Any discrepancy between a generated test case and a verifier's verdict is flagged for collective review and consensus. A task was only included in the final benchmark if the generator's ground truth perfectly matched the verifier's execution, thereby ensuring 100\% mathematical validity. The construction pipeline of ChomskyBench is shown in Figure \ref{fig:construction_pipline}.
This process yields an automated and scalable evaluation framework. For each task, we develop a programmatic generator capable of producing an infinite number of test instances on the fly for any specified complexity parameter $n$. An example of ChomskyBench is shown in Figure \ref{fig: example}. This enables detailed \textbf{asymptotic performance analysis}, revealing not just if a model fails, but precisely at what threshold of complexity its reasoning capabilities begin to break down.
\subsection{The Superiority and Core Contributions of ChomskyBench}
ChomskyBench offers four key advantages for evaluating the formal reasoning capabilities of LLMs:
\begin{enumerate}[label=\arabic*)]
    \item \textbf{A Solid Theoretical Foundation:} It is the first benchmark to systematically leverage the full Chomsky Hierarchy, including Type-0 (Recursively Enumerable) tasks, as its evaluation framework. Combined with its process-trace evaluation paradigm and symbolic verifiability, this moves evaluation beyond surface-level code functionality to a fundamental probe of a model's intrinsic computational power.
    \item \textbf{Unparalleled Diagnostic Power:} Its hierarchical design provides a high-resolution diagnostic tool. ChomskyBench yields not a single score, but a detailed "capability map" across the complexity spectrum. The model's performance decay curve, for instance, a sharp drop when moving from Type-2 to Type-1 tasks, provides a clear signal about its specific architectural limitations (e.g., an inability to handle multi-variable dependencies).
    \item \textbf{A Robust and Reliable Evaluation Standard:} The benchmark's reliance on deterministic algorithmic verifiers provides an unprecedented level of evaluation rigor. It eliminates the subjectivity of human scoring and the potential flakiness of unit-test-based evaluations, enabling an accurate, instantaneous, and infinitely scalable assessment. 
    \item \textbf{Resistance to Contamination:} The abstract and formal nature of these tasks makes it extremely unlikely that they appear verbatim in the web-scraped data used for pre-training. 
\end{enumerate}

\section{Experimental Design}
To systematically evaluate the formal reasoning capabilities of LLMs, we designed a comprehensive experiment guided by the ChomskyBench framework. Our empirical study addresses five research questions (RQs).
\begin{itemize}
    \item \textbf{RQ1: Hierarchical Performance Stratification.} How does the performance of SOTA LLMs vary across the different levels of the Chomsky Hierarchy (Type-3 to Type-0)? 
    \item \textbf{RQ2: Impact of Structural Complexity.} Within each hierarchical level, how does increasing the structural complexity parameter $n$ (e.g., string length, recursion depth) affect the reasoning performance of LLMs? This question investigates the "breaking point" of the models, identifying the thresholds at which their reasoning mechanisms begin to fail.
    \item \textbf{RQ3: Comparative Efficiency.} How does the computational efficiency (i.e., inference time) of LLMs on formal language tasks compare to that of their corresponding traditional, deterministic algorithms? 
    \item \textbf{RQ4: Influence of Test-time Scaling.} What is the impact of test-time scaling approaches, increasing sample size and extended reasoning length, on the ability of LLMs to solve formal reasoning tasks?
    \item \textbf{RQ5: Root Cause Analysis.} What are the primary failure modes that cause LLMs to fail on formal reasoning tasks? Do LLMs fail due to a lack of understanding of formal specifications, or due to execution errors during step-by-step reasoning?
\end{itemize}
\subsection{Studied LLMs}
To ensure a comprehensive and representative analysis of the latest released LLMs, we select 12 popular models covering a wide spectrum of recent open-source and closed-source models, reasoning and non-reasoning models, as well as general-purpose and code-specific models. The open-source models span a parameter range from 30B to 1T. Furthermore, for models within the same family, we include multiple versions with varying scales (e.g., Qwen3-30B-A3B and Qwen3-235B-A3B) to investigate the impact of parameter size on performance. For certain models (e.g., DeepSeek-V3.1), we evaluate both their reasoning and non-reasoning modes. As shown in Table~\ref{tab:studied_llms}, we summarize the studied LLMs along with their categories, model size(activated/total), training base, API price, context window(in/out), and data cutoff date.
\begin{table}[t]
\centering
\caption{Statistics of evaluated LLMs in our experiments. }
\label{tab:studied_llms}
\resizebox{\textwidth}{!}{
\begin{tabular}{lcccccc}
\toprule
Model & \makecell{Closed Source / \\ Open Source} & \makecell{Model Size} & \makecell{Training \\ Tokens} & \makecell{API Price} & \makecell{Context \\ Window} & \makecell{Data \\ Cutoff} \\
\midrule
deepseek v3.1 & Open & 37B/$\sim$671B & 14.8T & \$0.25/M In \$1/M Out & 128K & 07/2024 \\
gemini-2.5-pro & Closed & - & - & \$1.25/M In \$10/M Out & 1M/64K & 01/2025 \\
gemini-2.5-flash & Closed & - & - & \$0.3/M In \$2.5/M Out & 1M/64K & 01/2025 \\
gpt-5 & Closed & - & - & \$1.25/M In \$10/M Out & 272K/128K & 09/2024 \\
gpt-4o & Closed & - & - & \$2.5/M In \$10/M Out & 128K & 06/2024 \\
kimi-k2 & Open & 32B/$\sim$1T & 15.5T & \$0.38/M In \$1.52/M Out & 256K & 06/2024 \\
llama-4-maverick & Open & 17B/$\sim$400B & 40T & \$0.15/M In \$0.6/M Out & 1M & 08/2024 \\
o3 & Closed & - & - & \$2/M In \$8/M Out & 200K/100K & 05/2024 \\
qwen3-235b-a22b & Open & 22B/$\sim$235B & 36T & \$0.098/M In \$0.39/M Out & 262K & - \\
qwen3-30b-a3b & Open & 3.3B/30.5B & 36T & \$0.071/M In \$0.28/M Out & 262K & - \\
qwen3-coder-30b-a3b & Open & 3.3B/30.5B & 36T & \$0.071/M In \$0.28/M Out & 262K & - \\
sonnet-4 & Closed & - & - & \$3/M In \$15/M Out & 200K & 03/2025 \\
\bottomrule
\end{tabular}}
\end{table}
\subsection{Evaluation Metrics}
Each question consists of multiple sub-questions with answers in tuple form $A = (A_1, A_2, ..., A_n)$, and we employ \textbf{Accuracy (ACC):} Correct only if all sub-questions are answered correctly and \textbf{Pass Ratio (PR)} \cite{APPS, Codescore}: The average correctness rate across all sub-questions, to evaluate:  
\begin{equation*}
    ACC = \frac{1}{t}\sum_{i=1}^{t} \mathbb{I}[\text{all correct}] \quad \text{and} \quad PR = \frac{1}{t}\sum_{i=1}^{t} \frac{1}{p_i}\sum_{j=1}^{p_i} \mathbb{I}[\text{correct}_{i,j}].
\end{equation*}
\subsection{Implementation Details}
\label{sec:implementation}
We evaluated the LLMs listed in Table~\ref{tab:studied_llms} via the OpenRouter's API\footnote{https://openrouter.ai} in our experiments. For test-time scaling experiments, we used $temperature = 0.6$, $top_p = 0.95$, and $n = 32$ to sample multiple candidate responses; in all other experiments, we set $temperature = 0$ and $n = 1$ to obtain deterministic results. All other hyperparameters remained at their default values. Each experiment was carried out in a zero-shot setting with a structured system prompt designed to facilitate the automated extraction of model responses: ``You are a helpful AI assistant. Your goal is to solve the following problem and provide the final answer. The final answer must be enclosed within a `\textbackslash box\{\}' command. For example: `\textbackslash box\{Your Final Answer\}'.'' Throughout the experiments, we meticulously recorded the time required for each inference instance.

\section{Experimental Results}
\begin{table}[]
    \caption{Comparison of SOTA LLMs on automata and formal language tasks across different Chomsky hierarchy levels, which reports Accuracy (Acc) and PassRatio (PR) for both closed-source and open-source models, reflecting their comprehensive formal reasoning capabilities in regular, DCF, NCF, CS, and RE tasks.}
    \label{tab:sota_llms_performance}
    \resizebox{\textwidth}{!}{
    \begin{tabular}{lllllllllllll}
    \toprule
    \multirow{2}{*}{LLMs}                   & \multicolumn{2}{c}{Regular} & \multicolumn{2}{c}{DCF} & \multicolumn{2}{c}{NCF}  & \multicolumn{2}{c}{CS}  & \multicolumn{2}{c}{RE}  & \multicolumn{2}{c}{Avg} \\
    \cmidrule(l){2-3} \cmidrule(l){4-5} \cmidrule(l){6-7}\cmidrule(l){8-9}\cmidrule(l){10-11}\cmidrule(l){12-13}
    & Acc           & PR    & Acc        & PR  & Acc        & PR  & Acc        & PR  & Acc        & PR  & Acc        & PR  \\
    \midrule
    \textbf{Closed-source} {\tiny\faLock}                &               &               &            &            &            &            &            &            &            &            &            &            
    \\
    \textbf{o3}                    & \textbf{0.333}      & \textbf{0.619}          & \textbf{0.207}    & 0.491        & \textbf{0.286}    & \textbf{0.548}        & \textbf{0.250}    & \textbf{0.461}        & \textbf{0.217}    & 0.489        & \textbf{0.278}    & \textbf{0.563}        \\
    \textbf{gpt-5}                 & 0.250      & 0.569          & 0.069    & 0.457        & 0.190    & 0.405        & 0.214    & 0.435        & \textbf{0.217}    & 0.478        & 0.200    & 0.509        \\
    \textbf{gemini-2.5-pro}        & 0.250      & 0.500          & 0.103    & 0.500        & \textbf{0.286}    & 0.524        & 0.143    & 0.310        & 0.087    & 0.424        & 0.183    & 0.486        \\
    \textbf{gpt-4o}                & 0.125      & 0.290          & 0.172    & 0.483        & 0.048    & 0.393        & 0.036    & 0.336        & 0.043    & 0.413        & 0.096    & 0.418        \\
    \textbf{gemini-2.5-flash}      & 0.250      & 0.500          & 0.138    & 0.448        & 0.143    & 0.452        & 0.036    & 0.289        & 0.087    & 0.402        & 0.139    & 0.451        \\
    \textbf{sonnet-4}              & \textbf{0.333}      & 0.613          & \textbf{0.207}    & \textbf{0.543}        & 0.238    & 0.452        & 0.071    & 0.435        & 0.130    & \textbf{0.511}        & 0.209    & 0.556        \\
    \hdashline
    \textbf{Open-source} {\tiny \faKey}                    &               &               &            &            &            &            &            &            &            &            &            &            
    \\
    \textbf{kimi-k2}               & 0.292      & 0.521          & 0.172    & 0.440        & 0.238    & 0.488        & 0.107    & 0.372        & \textbf{0.174}    & 0.402        & 0.209    & 0.480        \\
    \textbf{llama-4-maverick}      & 0.167      & 0.420          & 0.069    & 0.460        & 0.000    & 0.381        & 0.036    & 0.384        & 0.043    & 0.304        & 0.070    & 0.428        \\
    \textbf{qwen3-235b-a22b}       & \textbf{0.417}      & \textbf{0.584}          & \textbf{0.207}    & \textbf{0.552}        & 0.190    & 0.488        & \textbf{0.143}    & \textbf{0.432}        & 0.130    & \textbf{0.402}        & \textbf{0.235}    & \textbf{0.536}        \\
    \textbf{qwen3-30b-a3b}         & 0.125      & 0.217          & 0.000    & 0.276        & 0.000    & 0.310        & 0.071    & 0.226        & 0.043    & 0.185        & 0.052    & 0.263        \\
    \textbf{qwen3-30b-a3b-coder}   & 0.083      & 0.355          & 0.034    & 0.313        & 0.048    & 0.298        & 0.036    & 0.381        & 0.000    & 0.272        & 0.043    & 0.355        \\
    \textbf{deepseek v3.1}       &  0.058 & 0.225 & 0.183 & 0.505 & 0.219 & 0.533 & 0.091 & 0.407 & 0.095 & 0.437 & 0.125 & 0.419  \\
    \bottomrule
    \end{tabular}}
\end{table}
\subsection{RQ1: Performance of SOTA LLMs on ChomskyBench}
\label{sec:rq1_performance}
To answer RQ1, we systematically evaluated a suite of SOTA LLMs on ChomskyBench. The results, presented in Table~\ref{tab:sota_llms_performance}, reveal a clear and consistent stratification of performance that directly correlates with the complexity of the Chomsky Hierarchy. Our analysis uncovers three key findings that map the theoretical limits of current models.
\paragraph{Performance Degrades with Increasing Grammatical Complexity}
The most prominent trend across all evaluated models is a sharp, monotonic decrease in performance as we ascend the Chomsky Hierarchy. While models demonstrate some capability on Regular and CF languages, their reasoning abilities falter significantly when confronted with the stricter rules of higher-order grammars. This is best exemplified by the top-performing model, o3, whose accuracy falls from 0.286 on NCF tasks to 0.250 on CS tasks, and further to 0.217 on RE tasks. This pattern strongly indicates that current LLM architectures struggle to manage the increasingly complex constraints, such as long-range dependencies and context-dependent rules, that define advanced formal languages.
\paragraph{A Decisive Performance Cliff Exists Between Context-Free and Context-Sensitive Languages}
Our analysis identifies a distinct performance ``cliff'' at the transition from Type-2 (Context-Free) to Type-1 (Context-Sensitive) languages. For the majority of models, accuracy on CS tasks drops substantially compared to CF tasks. For example, sonnet-4's accuracy plummets from 0.238 on NCF to 0.071 on CS, deepseek v3.1 drops from 0.219 on NCF to 0.091 on CS, and gemini-2.5-pro falls from 0.286 on NCF to 0.143 on CS. This cliff suggests a fundamental limitation: while LLMs may be able to approximate context-free structures through sophisticated pattern matching, they lack the core mechanisms to rigorously apply the context-dependent generative rules that are the hallmark of CS grammars. Even models with specialized reasoning capabilities, such as o3, show greater resilience but are still significantly impaired (0.250 Acc on CS), highlighting this as a deep-seated architectural challenge rather than a simple knowledge gap.
\paragraph{Deep Reasoning Enhances Resilience and Navigates Non-Determinism}
The deep reasoning model o3 is unequivocally the top performer in our study, achieving the highest average Accuracy (0.278) and PassRatio (0.563). This superior performance underscores the critical role of structured, step-by-step reasoning (e.g., long Chain-of-Thought) in tackling formal language tasks. More profoundly, o3 reveals a nuanced capability in handling non-determinism. While most models perform worse on NCF tasks than on Deterministic Context-Free (DCF) tasks (cf. llama-4-maverick: 0.069 Acc on DCF vs. 0.000 on NCF), o3 achieves a higher PassRatio on NCF (0.548) than on DCF (0.491). This counterintuitive result suggests that its reasoning process is effective at exploring the multiple valid derivation paths inherent in non-deterministic grammars. This enhances its ability to produce a valid reasoning trace (improving PassRatio), even if the final answer's correctness (Acc) remains constrained by the task's complexity.
\begin{boxK}
    \small \faIcon{pencil-alt} \textbf{Finding 1:}
    LLM performance is strictly stratified by the Chomsky Hierarchy, with a significant performance cliff at the transition from context-free to context-sensitive languages. Even the best models (e.g., o3) achieve only modest accuracy on CS/RE tasks, revealing fundamental limitations in processing high-order, recursive, and context-dependent structures.
\end{boxK}
\begin{figure}[t]
    \centering
    \begin{subfigure}[b]{0.48\textwidth}
        \centering
        \includegraphics[width=\textwidth]{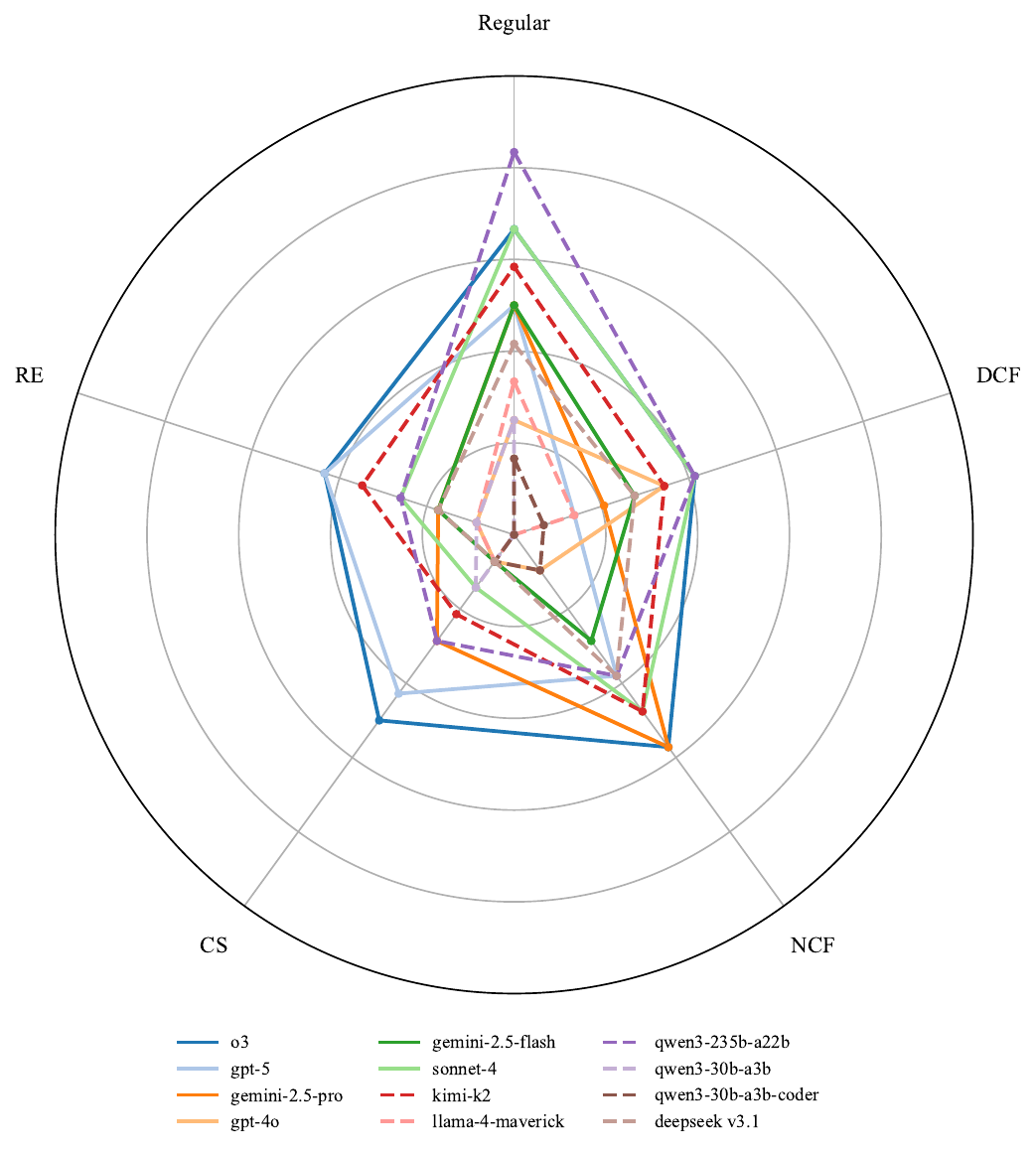}
        \caption{LLM Performance across the Chomsky Hierarchy}
        \label{fig:left}
    \end{subfigure}
    \hfill
    \begin{minipage}[b]{0.48\textwidth}
        \centering
        \begin{subfigure}{\textwidth}
            \centering
            \includegraphics[width=0.9\textwidth]{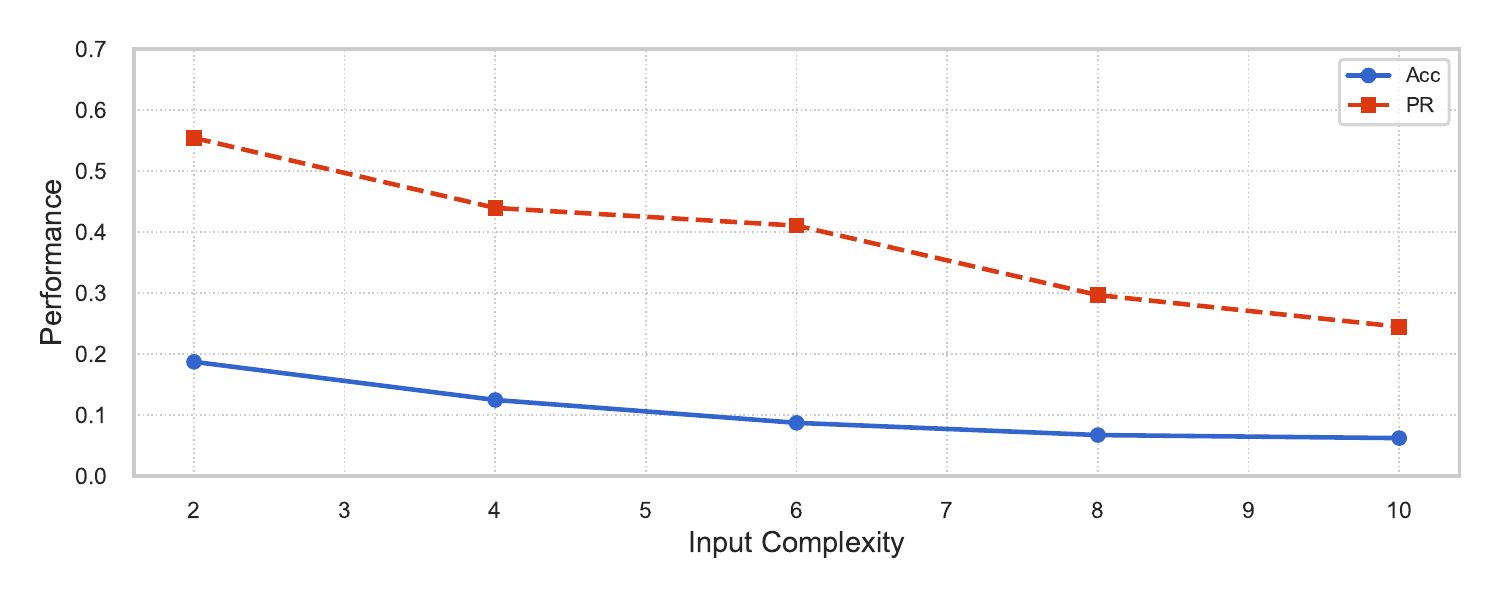}
            \caption{Impact of Input Complexity on LLM Performance}
            \label{fig:top_right}
        \end{subfigure}
        \begin{subfigure}{\textwidth}
            \centering
            \includegraphics[width=1\textwidth]{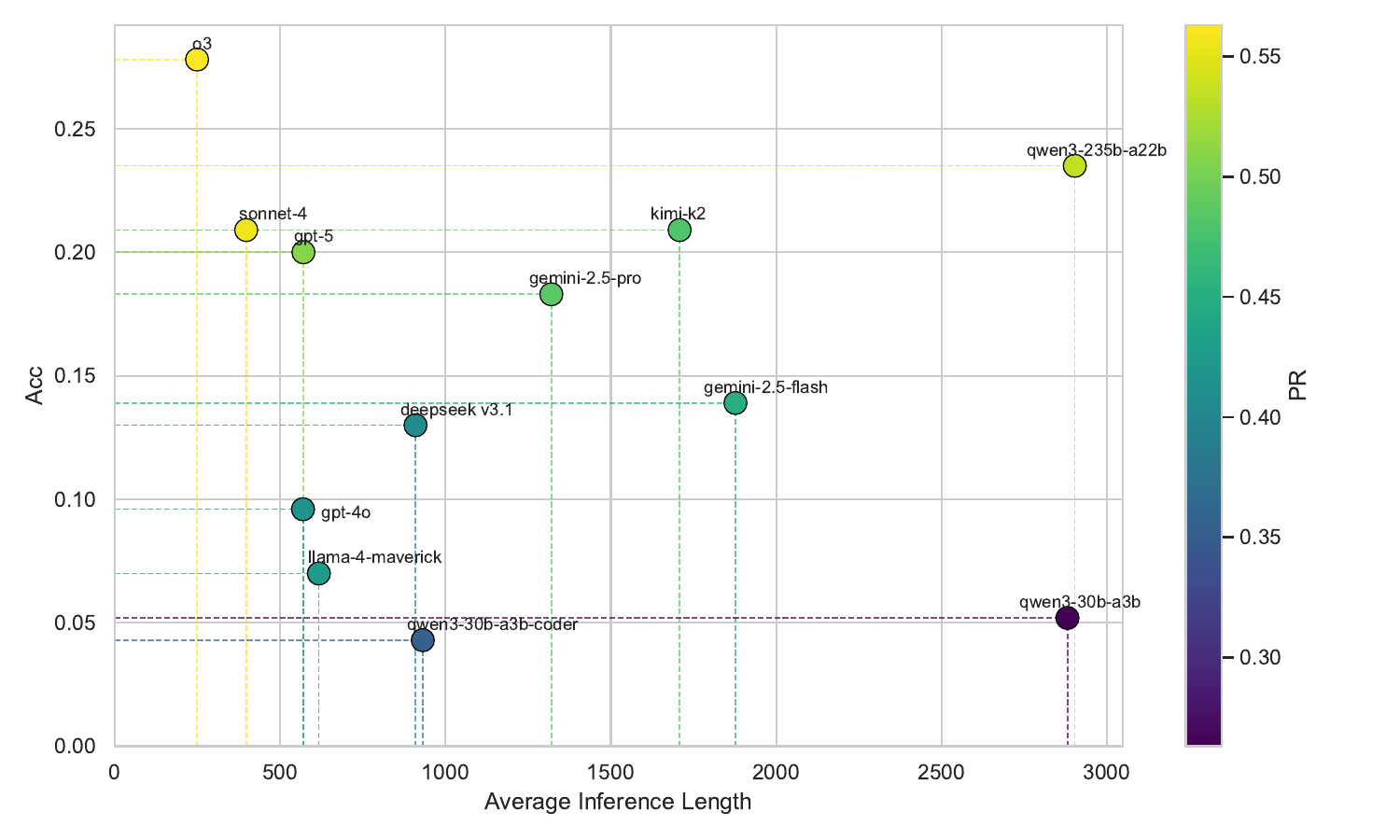}
            \caption{SOTA LLM's Performance and Inference Length}
            \label{fig:bottom_right}
        \end{subfigure}
    \end{minipage}
    \caption{A multi-faceted visualization of SOTA LLMs for formal reasoning performance on ChomskyBench. 
    }
    \label{fig:main}
\end{figure}
\subsection{RQ2: Affecting Factors for LLM Performance}
\label{sec:rq2}
To delve deeper into the how and why behind the performance stratification observed in RQ1, we investigate the dynamics of the models' inference processes. As visualized in Figure \ref{fig:left}, our results reveal a clear hierarchy: an LLM’s reasoning reliability systematically degrades as the formal language complexity increases, with a significant performance drop from simpler levels (e.g., Regular, DCF) to more complex ones (CS, RE). A key architectural feature of ChomskyBench is its designed scalability, allowing us to systematically vary input complexity (e.g., sequence length). This provides a unique lens through which to measure how the integrity of an LLM's reasoning holds up under increasing strain.
\paragraph{Performance is Inversely Correlated with Input Complexity}
As illustrated in Figure \ref{fig:top_right}, our analysis reveals a stark and unambiguous inverse relationship between input complexity and model performance. Across all tested models, both Accuracy (Acc) and PassRatio decline precipitously as the length of the input sequence increases. This trend is not a gradual tapering but a sharp drop, indicating that the reasoning mechanisms of current LLMs are not robust to scale. Their ability to maintain logical consistency and adhere to formal rules degrades rapidly when faced with longer problem instances, even for simpler language classes. This fragility at scale underscores a fundamental limitation in current architectures.
\paragraph{Reasoning Quality, Not Length, Determines Success}
Our findings expose a critical distinction between reasoning depth and reasoning quality, as illustrated in Figure \ref{fig:bottom_right}. This visualization maps each model's Average Inference Length against its Accuracy, revealing that longer reasoning chains do not inherently lead to better outcomes. In fact, for many models, the opposite is true. For instance, models like qwen3-235b-a22b produce some of the longest inference traces yet achieve only mediocre accuracy. Closer inspection reveals that this extended length often stems from low-quality, structurally repetitive reasoning loops \cite{Repetition,RethinkingRepetition}, where the model repeatedly attempts and fails the same logical steps without making substantive progress.
In contrast, top-performing reasoning LLMs like o3 exhibit a markedly different pattern. It achieves superior accuracy and PassRatio with a comparatively moderate inference length, indicating a more efficient and higher-quality reasoning process. Instead of brute-force exploration or repetition, its inference is more directed and logically sound, successfully navigating the problem's constraints without generating excessive, unproductive steps. This demonstrates conclusively that the quality and logical coherence of the reasoning trace, not its sheer length, are the primary determinants of success on complex formal tasks.
\begin{boxK}
\small \faIcon{pencil-alt} \textbf{ Finding 2:}
LLM performance sharply declines with increasing input complexity. Longer reasoning chains do not guarantee correctness, i.e., reasoning quality, not length, determines success.
\end{boxK}
\begin{table}[h!]
    \caption{Comparison of Program Execution Time and LLM Reasoning Time, and their respective Pearson correlations with PassRatio, i.e., $p_{e\&pr}$ and $p_{l\&pr}$, respectively.}
    \label{tab:code&llm}
    \resizebox{\textwidth}{!}{
    \begin{tabular}{@{}l *{15}{l}@{}}
    \toprule
    & \multicolumn{3}{c}{Regular} & \multicolumn{3}{c}{DCF} & \multicolumn{3}{c}{NCF} & \multicolumn{3}{c}{CS} & \multicolumn{3}{c}{RE} \\
    \cmidrule(l){2-4} \cmidrule(l){5-7} \cmidrule(l){8-10} \cmidrule(l){11-13} \cmidrule(l){14-16}
    & Time    & $p_{e\&pr}$ & $p_{l\&pr}$    & Time    & $p_{e\&pr}$ & $p_{l\&pr}$ & Time    & $p_{e\&pr}$ & $p_{l\&pr}$ & Time    & $p_{e\&pr}$ & $p_{l\&pr}$ & Time    & $p_{e\&pr}$ & $p_{l\&pr}$ \\
    \midrule
    \textbf{Program} & $<0.001$ & - & - & $<0.001$ & - & - & $<0.001$ & - & - & $<0.001$ & - & - & $<0.001$ & - & - \\
    \hdashline
    \textbf{Closed-source} {\tiny\faLock} \\
    \textbf{o3}               & 20.67  & 0.39  & -0.34 & 26.41  & 0.10  & -0.22 & 44.45  & -0.62 & -0.54 & 76.68  & -0.06 & -0.18 & 40.35  & -0.31 & -0.02 \\
    \textbf{gpt-5}            & 46.68  & -0.23 & -0.07 & 69.73  & 0.24  & 0.08  & 77.62  & -0.30 & -0.51 & 121.63 & 0.03  & -0.34 & 66.53  & -0.21 & -0.33 \\
    \textbf{gemini-2.5-pro}   & 48.85  & 0.38  & -0.09 & 54.49  & -0.27 & -0.23 & 58.76  & -0.54 & -0.60 & 105.44 & -0.24 & -0.44 & 74.50  & -0.21 & -0.48 \\
    \textbf{gpt-4o}           & 10.68  & 0.23  & -0.06 & 10.37  & -0.06 & -0.43 & 10.83  & -0.43 & -0.09 & 16.13  & -0.33 & -0.25 & 20.95  & -0.23 & -0.31 \\
    \textbf{gemini-2.5-flash} & 11.19  & -0.14 & -0.12 & 10.59  & -0.25 & -0.45 & 15.76  & -0.20 & -0.36 & 31.98  & -0.28 & -0.53 & 21.13  & -0.25 & -0.28 \\
    \textbf{sonnet-4}         & 12.24  & 0.19  & 0.19  & 12.68  & 0.34  & -0.13 & 14.31  & -0.60 & -0.56 & 17.24  & -0.31 & -0.28 & 15.65  & -0.20 & -0.24 \\
    \hdashline
    \textbf{Open-source} {\tiny \faKey} \\
    \textbf{kimi-k2}          & 88.00  & 0.14  & -0.06  & 152.84 & 0.02  & -0.08 & 342.21 & -0.24 & -0.61 & 357.06 & -0.31 & -0.44 & 240.38 & -0.17 & -0.37 \\
    \textbf{llama-4-maverick} & 15.30  & 0.06  & 0.42  & 12.17  & 0.37  & -0.20 & 14.17  & -0.52 & -0.12 & 15.26  & -0.27 & -0.12 & 13.27  & -0.19 & -0.39 \\
    \textbf{qwen3-235b-a22b}  & 81.72  & 0.11  & -0.10 & 100.06 & -0.26 & -0.40 & 112.17 & -0.14 & -0.19 & 166.70 & 0.11  & -0.36 & 132.82 & -0.11 & -0.60 \\
    \textbf{deepseek v3.1}    & 51.96  & 0.16  & -0.06 & 52.18  & -0.25 & -0.15 & 60.87  & -0.65 & -0.48 & 78.13  & -0.13 & -0.09 & 86.77  & -0.44 & 0.01  \\
    \bottomrule
    \end{tabular}}
\end{table}
\subsection{RQ3: Program Execution Time vs. LLM Reasoning Time }
To evaluate the practical utility of LLMs in formal software engineering tasks, we conduct a analysis of their reasoning time against the execution time of deterministic programs designed for the same tasks. The results, detailed in Table~\ref{tab:code&llm}, reveal a disparity in both efficiency and reliability.
\paragraph{A Gap in Efficiency and Reliability}
LLM reasoning is profoundly inefficient compared to programmatic execution. 
Conventional programs solve these formal language problems in under a millisecond ($<0.001s$) with perfect accuracy. In contrast, LLMs require seconds to minutes of inference time, making them tens of thousands to over 350,000 times slower. This efficiency gap, coupled with the fact that LLM outputs are often incorrect (as established in RQ1), renders their current application in high-reliability domains untenable. This finding reaffirms the necessity of traditional program analysis tools and formal methods software, highlighting that LLMs, in their present form, possess clear disadvantages in both reliability and efficiency that prevent them from replacing rigorous, algorithmic solutions.
\paragraph{Task Complexity Reveals a Fundamental Reasoning Deficit}
Our analysis of the Pearson correlation between program execution time and LLM PassRatio ($p_{e\&pr}$) reveals a fundamental deficit in LLM deductive reasoning. Program execution time serves as a direct proxy for the task's intrinsic complexity—a longer execution time implies more computational steps are required. The data shows a predominantly negative correlation, especially for more complex language types (NCF, CS, and RE). This indicates that as the number of required reasoning steps increases, an LLM’s likelihood of success decreases. This is a fatal flaw for formal reasoning, as it demonstrates that LLMs struggle to maintain logical coherence over extended deductive chains, a core requirement for any formal system.
\paragraph{Longer Inference Time Signifies Struggle, Not Success}
We find that simply allowing an LLM more time to "reason" is not a reliable strategy for improving performance. The Pearson correlation between LLM reasoning time and PassRatio ($p_{l\&pr}$) is also consistently negative across most models and task types. While one might hope that longer inference times correspond to deeper, more successful deliberation, our results show the opposite: extended reasoning time is more often a symptom of the model struggling, likely caught in the low-quality, repetitive reasoning loops identified in RQ2. This suggests that methods aiming to boost performance by simply increasing inference time (e.g., test-time scaling) are unstable and unlikely to overcome the core reasoning limitations of the models.
\begin{boxK}
\small \faIcon{pencil-alt} \textbf{ Finding 3:}
LLMs are 10,000-350,000x slower than conventional programs and far less reliable, underscoring the indispensability of traditional software tools. Negative correlations between task complexity and LLM success reveal fundamental deficits in sustained deductive reasoning.
\end{boxK}
\begin{table}[h!]
    \caption{Test-time Scaling of LLMs on ChomskyBench, including approaches of increasing sample size and extended reasoning length.}
    \label{tab:ttc}
    \resizebox{\textwidth}{!}{
    \begin{tabular}{lllllllllllll}
    \toprule
    \multirow{2}{*}{Approaches}                  & \multicolumn{2}{c}{Regular} & \multicolumn{2}{c}{DCF} & \multicolumn{2}{c}{NCF}  & \multicolumn{2}{c}{CS}  & \multicolumn{2}{c}{RE}  & \multicolumn{2}{c}{Avg} \\
    \cmidrule(l){2-3} \cmidrule(l){4-5} \cmidrule(l){6-7}\cmidrule(l){8-9}\cmidrule(l){10-11}\cmidrule(l){12-13}
    & Acc           & PR    & Acc        & PR  & Acc        & PR  & Acc        & PR  & Acc        & PR  & Acc        & PR  \\
    \midrule
    \textbf{deepseek v3.1} &    0.058    &    0.225    &    0.183    &    0.505    &    0.219    &    0.533    &    0.091    &    0.407    &    0.095    &    0.437    &    0.125    &    0.419        \\
    \hdashline
    \textbf{Increasing sample size}
    \\
    \textbf{Majority voting N=32}  &  0.045    &    0.212    &    0.231    &    0.548    &    0.353    &    0.662    &    0.148    &    0.472    &    0.130    &    0.467    &    0.174    &    0.467     \\
    \textbf{Best of N-Oracle N=2}  &  0.075    &    0.296    &    0.244    &    0.569    &    0.311    &    0.628    &    0.149    &    0.509    &    0.138    &    0.500    &    0.178    &    0.498 \\
    \textbf{Best of N-Oracle N=4}  &  0.092    &    0.370    &    0.297    &    0.615    &    0.403    &    0.707    &    0.224    &    0.597    &    0.179    &    0.549    &    0.233    &    0.564 \\
    \textbf{Best of N-Oracle N=8}  &  0.110    &    0.437    &    0.346    &    0.661    &    0.486    &    0.776    &    0.310    &    0.677    &    0.213    &    0.588    &    0.286    &    0.624 \\
    \textbf{Best of N-Oracle N=16} &  0.125    &    0.487    &    0.398    &    0.713    &    0.576    &    0.834    &    0.388    &    0.747    &    0.250    &    0.623    &    0.340    &    0.678 \\
    \textbf{Best of N-Oracle N=32} &  0.136    &    0.531    &    0.462    &    0.769    &    0.647    &    0.868    &    0.444    &    0.821    &    0.304    &    0.652    &    0.391    &    0.727 \\
    \hdashline
    \textbf{Extended reasoning length} 
    \\
    \textbf{deepseek v3.1 (thinking)}& 0.208    &    0.381    &    0.138    &    0.431    &    0.19    &    0.488    &    0.036    &    0.253    &    0.087    &    0.337    &    0.139    &    0.406 \\
    \bottomrule
    \end{tabular}}
\end{table}
\subsection{RQ4: Efficacy of Test-Time Scaling on ChomskyBench}
Having established the baseline performance and inherent limitations of SOTA LLMs in RQ1-RQ3, we now investigate whether these deficits can be overcome through test-time scaling. Specifically, we explore two common strategies: \textbf{increasing the sample size} (e.g., Best-of-N sampling) and \textbf{extending the reasoning length} (i.e., prompting the model for a longer, more detailed Chain-of-Thought). Our findings, presented in Table~\ref{tab:ttc}, reveal that while these techniques can provide performance boosts, they are ultimately constrained by the same hierarchical barriers, exposing the deep-seated nature of the models' limitations.
\paragraph{Diminishing Returns Across the Hierarchy}
Both scaling strategies demonstrably improve LLM performance, but these gains exhibit important patterns across the Chomsky Hierarchy. As shown in Table~\ref{tab:ttc}, a Best-of-32-Oracle approach significantly raises deepseek v3.1's average accuracy from 0.125 to 0.391. While all task categories benefit from increased sampling, the absolute performance on high-complexity tasks remains inadequate for practical applications. For instance, accuracy on Regular languages improves from 0.058 to 0.136, on DCF from 0.183 to 0.462, and on NCF from 0.219 to 0.647. Even CS and RE tasks show notable relative gains (CS: 0.091 to 0.444; RE: 0.095 to 0.304). However, despite these improvements, the peak accuracy on CS (0.444) and RE (0.304) tasks remains far below the reliability threshold required for safety-critical formal verification tasks. This pattern suggests that while test-time scaling can amplify latent capabilities, the absolute performance ceiling on high-complexity formal reasoning remains a fundamental barrier.
\paragraph{The Efficiency of Quality over Quantity}
When comparing the two scaling methods, our results indicate that prompting for a single, higher-quality reasoning trace is a more resource-efficient strategy than generating numerous samples. The `deepseek v3.1 (thinking)` model, which was prompted for a longer CoT, achieved an average accuracy of 0.139. This is a modest improvement over the baseline (0.125) and is achieved at a fraction of the computational cost of the Best-of-N Oracle methods that yield comparable gains. For instance, achieving a similar level of performance requires at least N=4 samples, a 4x increase in inference cost. This aligns with our conclusion in RQ2: the quality and logical coherence of a reasoning path are more critical than the sheer quantity of attempts. Encouraging deeper, more structured deliberation in a single pass appears to be a more direct route to improving formal reasoning than brute-force sampling.
\paragraph{Fundamental Limits Persist}
Perhaps the most crucial insight from this analysis is what test-time scaling \textit{fails} to accomplish \textit{efficiently}. Even with a significant computational budget---generating and evaluating 32 distinct solutions for each problem---the LLM's performance remains inadequate for high-stakes applications. The Best-of-32-Oracle achieves only 0.462 on DCF (the highest), 0.647 on NCF, 0.304 on RE, 0.444 on CS, and merely 0.136 on Regular tasks---all far from the near-perfect reliability required for safety-critical formal verification.
Importantly, the scaling trend reveals both promise and concern. The accuracy gains per doubling of N remain remarkably consistent ($\sim$0.05 per doubling), with no hard performance ceiling observed. However, extrapolating this trend, achieving $>$90\% accuracy would require extremely large sample sizes (estimated N $>$ 10,000), incurring prohibitive computational costs. Moreover, the Best-of-N-Oracle method assumes access to a perfect verifier, that is, an unrealistic assumption since such oracles are unavailable in practice. This exposes a critical gap: while LLMs can generate correct solutions with sufficient sampling, they lack the intrinsic capability to reliably \textit{verify} their own outputs. Thus, while the absence of a hard ceiling leaves room for future improvements through algorithmic advances, current approaches face a severe \textbf{efficiency barrier}, i.e., the computational cost required to achieve practical reliability through scaling alone is impractical for real-world deployment.
\paragraph{Data Contamination Analysis}
To ensure the validity of our experimental results, we conduct a data contamination analysis using the CDD method~\cite{Data-Contamination}, following its original settings. We compute Peak and CR metrics for each Chomsky Hierarchy level. As shown in Table~\ref{tab:cdd}, the results indicate negligible data contamination for ChomskyBench, where all CR values are 0, and Peak values are near zero across all Chomsky level. The observed performance patterns genuinely reflect the models' formal reasoning capabilities rather than memorization artifacts.
\begin{table}[h]
\centering
\caption{Data Contamination Detection  CDD~\cite{Data-Contamination}, where Peak denotes the average peakedness of output distribution, and CR denotes the ratio of contaminated tasks. The lower value is better.}
\label{tab:cdd}
\small
\begin{tabular}{l cc cc cc cc cc cc}
\toprule
\multirow{2}{*}{\textbf{Model}} & \multicolumn{2}{c}{\textbf{Regular}} & \multicolumn{2}{c}{\textbf{DCF}} & \multicolumn{2}{c}{\textbf{NCF}} & \multicolumn{2}{c}{\textbf{CS}} & \multicolumn{2}{c}{\textbf{RE}} & \multicolumn{2}{c}{\textbf{Avg}} \\
\cmidrule(l){2-3} \cmidrule(l){4-5} \cmidrule(l){6-7} \cmidrule(l){8-9} \cmidrule(l){10-11} \cmidrule(l){12-13}
& Peak & CR & Peak & CR & Peak & CR & Peak & CR & Peak & CR & Peak & CR \\
\midrule
deepseek v3.1 & 0 & 0 & 0 & 0 & 0.015 & 0 & 0.015 & 0 & 0 & 0 & 0.006 & 0 \\
\bottomrule
\end{tabular}
\end{table}
\begin{boxK}
\small \faIcon{pencil-alt} \textbf{Finding 4:}
Test-time scaling improves performance consistently ($\sim$0.05 accuracy per doubling of N) with no hard ceiling observed. However, achieving practical reliability ($>$90\%) would require prohibitive computational costs (N $>$ 10,000) and assumes unrealistic oracle access, revealing a severe efficiency barrier rather than an absolute capability limit.
\end{boxK}
\subsection{RQ5: What Causes LLM Failures in Formal Reasoning?}
\label{sec:rq5_error_analysis}
To move beyond aggregate performance metrics and provide actionable insights, we conduct a systematic error analysis to diagnose the root causes of LLM failures on formal reasoning tasks.
\paragraph{LLMs Understand the Task but Fail in Execution}
A critical question is whether LLMs fail because they do not understand the formal language representation, or because they understand but cannot execute the required reasoning. To investigate this, we employ an LLM-as-a-Judge analysis using DeepSeek-V3.1 to evaluate error cases across three dimensions: (1) \textbf{Automata Terminology:} whether the model correctly interprets states, transitions, and acceptance conditions; (2) \textbf{Transition Rules:} whether the model correctly applies the formal rules; and (3) \textbf{Output Format:} whether the response follows the required structure.
\begin{table}[h]
\centering
\caption{LLM-as-Judge Analysis of Error Cases: Understanding vs. Execution, where higher value means understanding better.}
\label{tab:error_analysis}
\small
\begin{tabular}{lccc}
\toprule
\textbf{Model} & \textbf{Terminology (1-5)} & \textbf{Transition Rules (1-5)} & \textbf{Output Format (1-5)} \\
\midrule
gpt-4o & 4.09 & 3.21 & 4.29 \\
sonnet-4 & 4.09 & 3.53 & 4.52 \\
gemini-2.5-pro & 3.65 & 3.49 & 4.14 \\
deepseek-v3.1 & 4.26 & 3.77 & 4.48 \\
kimi-k2 & 4.30 & 3.83 & 4.29 \\
llama-4-maverick & 3.76 & 2.96 & 3.65 \\
qwen3-235b-a22b & 4.38 & 3.98 & 4.35 \\
\bottomrule
\end{tabular}
\end{table}
As shown in Table~\ref{tab:error_analysis}, high-performing LLMs consistently score above 4.0 on Automata Terminology and Output Format, indicating strong understanding of the task specification and formatting requirements. However, scores on Transition Rules are notably lower (ranging from 2.96 to 3.98), revealing that the primary failure mode is not comprehension but \textit{execution}---the models understand the formal rules but struggle to correctly simulate their step-by-step application.
\paragraph{Taxonomy of Failure Modes}
Our qualitative analysis of error cases reveals three primary failure modes that become increasingly prevalent at higher Chomsky Hierarchy levels:
\begin{itemize}
    \item \textbf{State Tracking Collapse:} The model loses track of the current automaton state during long execution traces, leading to incorrect transitions or premature termination.
    \item \textbf{Recursion Depth Limitations:} For deeply nested structures (e.g., $a^nb^n$ with large $n$), the model fails to maintain the implicit ``stack'' required to match opening and closing symbols.
    \item \textbf{Long-Range Dependency Failure:} In context-sensitive tasks requiring synchronization of multiple counts (e.g., $a^nb^nc^n$), the model cannot reliably maintain and correlate independent counters across the sequence.
\end{itemize}
\begin{boxK}
\small \faIcon{pencil-alt} \textbf{Finding 5:}
LLMs understand formal specifications but fail in execution. Primary failure modes, i.e., state tracking collapse, recursion depth limitations, and long-range dependency failures, reveal that current architectures lack robust mechanisms for maintaining symbolic state during extended reasoning.
\end{boxK}

\section{Implications}
Our study is the first to systematically evaluate the formal reasoning capabilities of LLMs under the Chomsky Hierarchy, filling the gap left by existing benchmarks that lack a computational and complexity-oriented perspective. Based on the findings throughout our experiments, we summarize the following implications.
\paragraph{Mapping to Real-World Software Engineering Tasks}
To bridge the gap between our abstract formal language tasks and practical software engineering, we provide a mapping between Chomsky Hierarchy levels and real-world SE applications:
\begin{itemize}
    \item \textbf{Regular (Type-3):} Lexical analysis, token recognition, simple pattern matching in log parsing, and regular expression validation. LLMs show reasonable competence here.
    \item \textbf{Context-Free (Type-2):} Syntax parsing, code formatting, bracket matching, and basic AST construction. LLMs exhibit moderate capability but degrade with nesting depth.
    \item \textbf{Context-Sensitive (Type-1):} Semantic analysis (e.g., type checking, scope resolution), variable declaration-before-use validation, and cross-serial dependencies in protocol verification. LLMs show significant limitations here.
    \item \textbf{Recursively Enumerable (Type-0):} General program analysis, termination analysis, and full formal verification. LLMs are fundamentally unreliable for these tasks.
\end{itemize}
This mapping provides practitioners with a principled framework for assessing risk when deploying LLMs in software engineering pipelines.
\paragraph{Implications for researchers}
The experimental results reveal significant efficiency barriers for current LLMs in formal language tasks: as grammatical complexity increases, model performance declines systematically. While test-time scaling shows consistent improvements with no hard ceiling observed, achieving practical reliability would require prohibitive computational costs---our analysis suggests that reaching $>$90\% accuracy may require $>$10,000 samples per problem. This finding suggests that bridging the gap to safety-critical formal verification requires not just more computation, but fundamentally more efficient approaches. For future research, this highlights the necessity of moving beyond the “bigger-is-better” paradigm and exploring new model architectures that can better capture hierarchical reasoning patterns, thereby fundamentally enhancing formal reasoning capabilities.
\paragraph{Implications for practitioners}
Our results show that LLMs are inefficient in formal tasks and often fall into lengthy but unproductive reasoning when facing complex problems. This implies that, in real-world applications, LLMs are not suitable as standalone formal reasoning engines. Instead, practitioners should employ them as complementary tools for heuristic exploration in combination with traditional algorithmic methods, rather than as replacements. At the same time, deployment should take into account the limitations of LLMs in complex reasoning, avoiding over-reliance on “longer thinking time” to compensate for performance gaps, and instead improving effectiveness through task decomposition, external tool integration, and hybrid workflows.

\section{Threats to Validity}
\label{sec:threats}
\paragraph{Construct Validity} Construct validity concerns whether ChomskyBench and its associated metrics accurately measure the core concept of formal reasoning. A primary consideration is that our abstract, symbolic tasks may not fully represent real-world software engineering scenarios. However, this abstraction is a deliberate and necessary design choice, not a limitation. By stripping away semantic context and language-specific syntax, we isolate the foundational computational primitives, such as recursion, state management, and long-range dependency tracking, that are prerequisites for any complex formal task. 
\paragraph{Internal Validity}
Internal validity addresses potential confounding factors that could offer alternative explanations for our results. A key threat is the potential confounding of task complexity with input sequence length, as tasks higher on the Chomsky Hierarchy often require longer strings. To decouple these variables, our experimental design featured overlapping length distributions across all hierarchy levels. Our analysis confirms that the performance degradation between hierarchical levels is significantly steeper than that observed from increasing length within a single level, isolating computational complexity as the dominant independent variable. Moreover, we ensure the integrity of ChomskyBench. All tasks were derived from canonical definitions in the Theory of Computation and verified by multiple domain experts. To mitigate the risk of data contamination \cite{Data-Contamination, bubeck2023sparks}, all test instances are generated by humans with pure symbols.
\paragraph{External Validity}
External validity concerns the generalizability of our findings to other LLMs and real-world software engineering tasks. Our study included a diverse set of SOTA LLMs, varying in architecture and size. The consistent results observed across all models suggest that our findings are not artifacts of a specific model but rather indicative of a fundamental limitation of current LLMs. Crucially, while our tasks are abstract, our conclusions are grounded in the principles of computation theory, allowing for principled generalization. 

\section{Conclusion and Future Work}
We introduced ChomskyBench, the first comprehensive benchmark designed to evaluate the formal reasoning abilities of LLMs through the full Chomsky Hierarchy, combining process-trace evaluation via natural language with deterministic symbolic verifiability. We demonstrate that the performance of current SOTA LLMs exhibits clear stratification across the hierarchy. While test-time scaling methods show consistent improvements with no hard performance ceiling observed, achieving practical reliability (e.g., $>$90\% accuracy) would require prohibitive computational costs (estimated N $>$ 10,000 samples) and unrealistic oracle assumptions. This reveals a severe efficiency barrier rather than an absolute capability limit, but one that nonetheless renders current approaches unsuitable for strict formal reasoning tasks. 
Future work includes developing model architectures to improve the level of formal reasoning in LLMs, and exploring hybrid neural–symbolic approaches to tackle software engineering tasks, where current approaches (e.g., Kimina-Prover, LeanDojo) cannot even work in our tasks. 

\bibliographystyle{ACM-Reference-Format}
\bibliography{reference}
\end{document}